\documentclass[runningheads]{llncs}
\usepackage[T1]{fontenc}
\usepackage{times}  
\usepackage{helvet}  
\usepackage{courier}  
\usepackage[hyphens]{url}  
\usepackage{graphicx} 
\usepackage{subfigure}
\usepackage[caption=false,font=footnotesize]{subfig}
\usepackage{times}
\usepackage{helvet}
\usepackage{courier}
\usepackage{xcolor}
\usepackage{caption} 
\usepackage{algorithm}
\usepackage{algpseudocode}
\usepackage[nolist]{acronym}
\usepackage{amsmath,amsfonts,amssymb}
\usepackage{textcomp}
\usepackage{psfrag}
\usepackage{multimedia}
\usepackage{fancybox}
\usepackage{mathtools}
\usepackage{pgfplots}    
\usepackage{pgf}         
\pgfplotsset{compat=1.18}      

\usepackage{adjustbox}
\usepackage{tikz}
\usetikzlibrary{calc}
\usetikzlibrary{shapes.geometric}
\usepackage{booktabs}
\usepackage{siunitx}  
\usepackage{algcompatible}
\usepackage{newfloat}
\usepackage{listings}
\DeclareCaptionStyle{ruled}{labelfont=normalfont,labelsep=colon,strut=off} 
\floatstyle{ruled}
\newfloat{listing}{tb}{lst}{}
\floatname{listing}{Listing}
\begin{document}
\begin{acronym}
    \acro{sdm}[SDM]{Sequential Decision Making}
   \acro{mdp}[MDP]{Markov Decision Process}
   \acro{rl}[RL]{Reinforcement Learning}
   \acro{sl}[SL]{Supervised Learning}
   \acro{ssl}[SSL]{Self-Supervised Learning}
   \acro{dp}[DP]{Dynamic Programming}
   \acro{sdp}[SDP]{Stochastic Dynamic Programming}
   \acro{adp}[ADP]{Approximate Dynamic Programming}
   \acro{sp}[SP]{Stochastic Programming}
   \acro{lp}[LP]{Linear Programming}
   \acro{mpc}[MPC]{Model Predictive Control}
   \acro{empc}[EMPC]{Economic Model Predictive Control}
   \acro{td}[TD]{Temporal Difference}
   \acro{mbrl}[MBRL]{Model-based Reinforcement Learning}
   \acro{dt}[DT]{Digital Twin}
   \acro{ml}[ML]{Machine Learning}
   \acro{lstm}[LSTM]{Long Short-Term Memory}
   \acro{rnn}[RNN]{Recurrent Neural Network }
   \acro{dnn}[DNN]{Deep Neural Network }
   \acro{dnns}[DNNs]{Deep Neural Networks}
   \acro{gp}[GP]{Gaussian Processes}
   \acro{ai}[AI]{Artificial Intelligence}
   \acro{mle}[MLE]{Maximum Likelihood Estimation}
   \acro{lqr}[LQR]{Linear Quadratic Regulator}
   \acro{gpu}[GPU]{Graphics processing unit}
   \acro{tpu}[TPU]{Tensor processing unit}
   \acro{gans}[GANs]{Generative Adversarial Networks}
   \acro{cnns}[CNNs]{Convolutional Neural Networks}
   \acro{ls}[LS]{Least Squares}
   \acro{mse}[MSE]{Mean Squared Error}
    \acro{map}[MAP]{Maximum A Posteriori}
    \acro{gmm}[GMM]{Gaussian Mixture Model}
    \acro{hmm}[HMM]{Hidden Markov Model}
    \acro{elbo}[ELBO]{Evidence Lower Bound}
    \acro{vae}[VAE]{Variational Autoencoder}
    \acro{gan}[GAN]{Generative Adversarial Network}
    \acro{ode}[ODE]{Ordinary Differential Equations}
    \acro{ude}[UDE]{Universal Differential Equations} 
    \acro{ham}[HAM]{Hybrid Analysis and Modeling} 
    \acro{vi}[VI]{Variational Inference} 
    \acro{em}[EM]{Expectation-Maximization} 
    \acro{lstdq}[LSTDQ]{Least Square Temporal Difference Q-learning} 
    \acro{rldp}[RLDP]{Reinforcement Learning of Decision Processes}
    \acro{or}[OR]{Operations Research}
    \acro{bnb}[B\&B]{Branch and Bound}
    \acro{co}[CO]{Combinatorial Optimization}
    \acro{milp}[MILP]{Mixed Integer Linear Program}
    \acro{dfl}[DFL]{Decision Focused Learning}
    \acro{pfl}[PFL]{Prediction Focused Learning}
    \acro{spg}[SPG]{Stochastic Policy Gradient}
    \acro{knn}[KNN]{K-Nearest Neighbor}
    \acro{nn}[NN]{Neural Network}
    \acro{gae}[GAE]{Generalized Advantage Estimation}
    \acro{nns}[NNS]{Nearest-Neighbor Sampling}
    \acro{knns}[KNNS]{K-Nearest-Neighbor Sampling}
    \acro{kkt}[KKT]{Karush–Kuhn–Tucker}
    \acro{sac}[SAC]{Soft Actor–Critic}
    \acro{ppo}[PPO]{Proximal Policy Optimization}
\end{acronym}

\title{CORL: A Reinforcement Learning Framework for Combinatorial Optimization}
\title{CORL: Reinforcement Learning of MILP Policies for \\Combinatorial Sequential Decision Making}
\title{\textsc{corl}: Reinforcement Learning of MILPs with Branch-and-Bound Policies}
\title{\textsc{corl}: Reinforcement Learning of MILP Policies \\Solved via Branch‐and‐Bound}

\author{Akhil S Anand\inst{1} \and
Elias Aarekol\inst{1} \and
Martin Mziray Dalseg\inst{1} \and 
Magnus Stålhane\inst{1} \and 
Sebastien Gros\inst{1}}
\authorrunning{Akhil S Anand et al.}
%
\institute{Norwegian University of Science and Technology (NTNU), Trondheim, Norway
\email{akhil.s.anand@ntnu.no}}
\maketitle              


\begin{abstract}
Combinatorial sequential decision-making problems are typically modeled as mixed‐integer linear programs (MILPs) and solved via branch‐and‐bound (B\&B) algorithms. The inherent difficulty of modeling MILPs that accurately represent stochastic real-world problems leads to suboptimal performance in the real world. Recently, machine‐learning methods have been applied to build MILP models for decision quality rather than how accurately they model the real-world problem. However, these approaches typically rely on supervised learning, assume access to true optimal decisions, and use surrogates for the MILP gradients. In this work, we introduce a proof‐of‐concept \textsc{corl} framework that end-to-end fine‐tunes an MILP scheme using reinforcement learning (RL) on real‐world data to maximize its operational performance. We enable this by casting an MILP solved by B\&B as a differentiable stochastic policy compatible with RL. We validate the \textsc{corl} method in a simple illustrative combinatorial sequential decision-making example.
\end{abstract}

\section{Introduction}
\ac{co} is widely used for decision-making in \ac{or}, with applications spanning from production scheduling and supply‐chain design to vehicle routing and portfolio management~\cite{korte2008combinatorial}.  A broad class of \ac{co} problems can be formulated as \acp{milp} and solved using the \ac{bnb} algorithm. The standard model‐based decision‐making workflow constructs an \ac{milp} model that closely approximates the real‐world problem and then solves it.  Recently, machine learning has been integrated to improve the predictive accuracy of \acp{milp} models as a proxy to improving their real‐world performance, termed as \ac{pfl}~\cite{eiselt2022operations,petropoulos2024operational,mandi2022decision}. However, in a model-based decision-making scheme for a stochastic real-world environment, the conditions for optimality are distinct from predictive accuracy~\cite{anand2025all,anand2024optimality}. This has been identified as the \emph{objective mismatch} issue within \ac{mbrl}~\cite{wei_unified_2024,grimm2021proper} and has been observed empirically~\cite{pmlr-v162-hansen22a,farahmand2017value,schrittwieser2020mastering,difftori,voelcker2022value}. 

In principle, \ac{co} problems can also be solved optimally using \ac{rl} directly without modeling them as \acp{milp}~\cite{bello2016neural}. However, such model-free \ac{rl} policies lack guarantees for constraint satisfaction, explainability, struggle to generalize reliably beyond the training distribution, and require a large amount of real-world interactions, reducing their practical applicability, unlike \acp{milp}, which offer these properties by construction. Alternatively, the \ac{dfl} framework uses machine learning to directly fine-tune \acp{milp} for decision quality, thereby preserving the \ac{milp} structure while improving performance through learning~\cite{mandi2024decision}. \ac{dfl} methods prioritize performance over predictive accuracy of the \ac{milp} models. \ac{rl} naturally supports this approach, offering a way to learn or fine-tune model-based decision policies (e.g., \ac{mpc} or \ac{mbrl}) directly for real-world performance rather than learning a black-box neural network policy~\cite{anand2025closing,amos_differentiable_2020,reiter2025synthesis}. 

An \ac{milp} scheme inherently defines a deterministic, non-differentiable model-based decision policy by mapping states to the integer decision that minimizes a given objective. By recasting this \ac{milp} policy as a parameterized policy within the \ac{rl} framework, one can make use of policy-gradient \ac{rl} methods to adapt the \ac{milp} policy for improved real-world performance. Policy-gradient algorithms require the policy to be both stochastic and differentiable, properties that a standard \ac{milp}-based decision scheme lacks. Currently, \ac{dfl} methods try to approximate the gradient of an \ac{milp} using different surrogates; and in addition, they rely on an oracle providing the true value of the objective function, making them not suitable for \ac{rl}. 

We propose the \textsc{corl} framework by converting a discrete \ac{milp}+\ac{bnb} policy into a differentiable stochastic policy that is compatible with \ac{rl}. \textsc{corl} provides the necessary tools to integrate this \ac{milp} policy with the stochastic policy gradient \ac{rl} methods. \textsc{corl}  enables end‐to‐end fine‐tuning of \ac{milp} parameters directly from data to maximize its real-world decision performance while preserving its structure. \textsc{corl} approach falls broadly under the umbrella of emerging \ac{dfl} methodologies within combinatorial decision-making.

\paragraph{Organization:} The rest of the paper is organized as follows:  Section~\ref{sec:realted_works} reviews the related works; Section~\ref{sec:background} provides the necessary background; Section~\ref{sec:method} introduces the \textsc{corl} framework; Section~\ref{sec:examples} presents an illustrative example; Section~\ref{sec:discussion} discusses the findings, limitations and future work and Section~\ref{sec:conclusions} provides the conclusions.

\section{Related Works}\label{sec:realted_works}
\ac{dfl} is a major class of learning-based \ac{co} methods that seeks to improve the performance of decisions by training the \ac{co} in an end-to-end fashion. Like \ac{pfl}, \ac{dfl} first predicts parameters for a \ac{co} instance whose solution is the decision.  Instead of minimizing parameter error, \ac{dfl} minimizes a task loss most commonly the \emph{regret}~\cite{kotary2021end,mandi2024decision,sadana2025survey}. \ac{dfl} depends on proxy gradients of the \ac{milp} layer, which existing methods obtain via three main strategies~\cite{mandi2024decision}. First class, the analytical smoothing approach relaxes the \ac{milp} to an \ac{lp}, then smoothes the \ac{lp} solution map to obtain approximate gradients~\cite{ferber2020mipaal,wilder2019melding}.  The second class of the methods models the decision variable as a stochastic variable and estimates the sensitivity by perturbing its parameters~\cite{poganvcic2019differentiation,berthet2020learning}. The third class optimizes a surrogate objective (e.g.\ SPO+), avoiding differentiation through the solver \cite{elmachtoub2022smart}. Unlike \ac{dfl} methods that rely on proxy gradients through either of these methods, our approach uses the \ac{bnb} tree to define a softmax‐based stochastic policy that is directly differentiable, avoiding the need to find a suitable proxy. Although the \ac{dfl} methods enable regret minimization, they depend on knowing the true \ac{milp} parameters and their gradients. Therefore, \ac{dfl} methods could fail when data is noisy or the true parameters are not known, limiting their applicability in real‐world settings~\cite{mandi2024decision}. They also pose the added complexity of backpropagating through the argmin operator~\cite{mandi2024decision}.  \textsc{corl} avoids these issues by approximating the \ac{milp} as a differentiable stochastic policy, allowing end-to-end \ac{rl} of a general \ac{milp} scheme to maximize its real-world performance.

\ac{rl} has been used in \ac{co} primarily to enhance solver performance or solution quality under a given model~\cite{mazyavkina2021reinforcement,qi2021smart}. For \acp{milp}, early work focused on learning branching rules within \ac{bnb}~\cite{khalil2016learning}, while others trained \ac{rl} agents to construct feasible solutions directly~\cite{ma2019combinatorial}. 
However, tuning an existing \ac{milp} model, i.e., adjusting its parameters or heuristics to optimize real‐world decision performance via \ac{rl} remains an open challenge. In a more closely related work, mixed‐integer \ac{mpc} has been used as a policy for \ac{rl}~\cite{gros2020reinforcement}. The proposed \textsc{corl} approach is more general as it exploits the properties of the \ac{bnb} solver and allows for all leaf nodes to enter the policy, given a uniform sampling.  Xu \emph{et al.}\ embedded a modified DQN into an \ac{milp} to select feasible integer decisions, but this approach is tied to the black-box nature of  DQN and cannot fine‐tune a pre‐trained \ac{milp} model~\cite{xu2025reinforcement}. In another related work~\cite{lee2024rl}, a Q‐function estimate is applied to an \ac{milp} model to fine‐tune decision performance similar to what we presented in our examples. However, \textsc{corl} is more general, does not require a Q-function estimate, and can be applied to any \ac{milp} model.

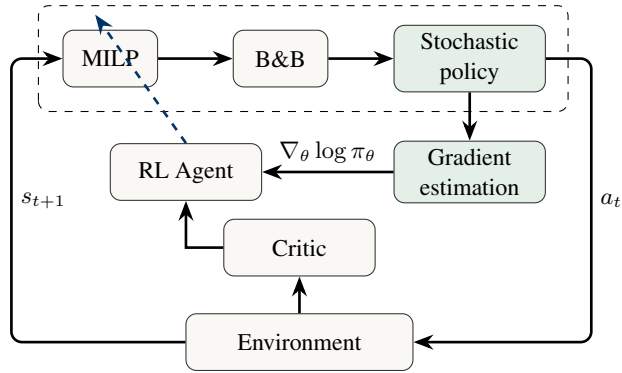
\begin{figure}[!t]
  \centering
  \resizebox{0.8\columnwidth}{!}{%
      \usetikzlibrary{fit,shapes,arrows.meta,positioning,calc}
  \usetikzlibrary{backgrounds}

\definecolor{Actor}{RGB}{255,255,255}   
\definecolor{MILP}{RGB}{250,248,245}
\definecolor{BNB}{RGB}{250,248,245}
\definecolor{Policy}{RGB}{228, 239, 231}    
\definecolor{Grad}{RGB}{228, 239, 231}      
\definecolor{RLAgent}{RGB}{250,248,245}
\definecolor{critic}{RGB}{250,248,245}
\definecolor{env}{RGB}{250,248,245}

  \begin{tikzpicture}[
      >=Stealth,
      arrow/.style={->, line width=1pt, black},
      actorbg/.style={
        draw=black, fill=Actor, rounded corners, inner sep=4pt},
      milpblock/.style={
        draw=black!80, fill=MILP, rounded corners,
        minimum width=1.25cm, minimum height=0.75cm},
      bnbblock/.style={
        draw=black!80, fill=BNB, rounded corners,
        minimum width=1.25cm, minimum height=0.75cm},
      policyblock/.style={
        draw=black!80, fill=Policy, rounded corners,
        minimum width=2cm, minimum height=0.75cm, align=center},
      gradblock/.style={
        draw=black!80, fill=Grad, rounded corners,
        minimum width=2cm, minimum height=0.75cm, align=center},
      rlblock/.style={
        draw=black!80, fill=RLAgent, rounded corners,
        minimum width=2cm, minimum height=0.75cm},
      criticblock/.style={
        draw=black!80, fill=critic, rounded corners,
        minimum width=2cm, minimum height=0.75cm},
      simblock/.style={
        draw=black!80, fill=env, rounded corners,
        minimum width=3cm, minimum height=0.75cm}
    ]

    \draw[actorbg, dashed] (0.55,4.80) rectangle (7.55,6.20);

    \node[milpblock]   (milp)   at (1.5,5.50) {MILP};
    \node[bnbblock]    (bnb)    at (3.75,5.50)   {B\&B};
    \node[policyblock] (policy) at (6.25,5.50) {Stochastic\\ policy};

    \node[gradblock]   (grad)   at (6.25,4.00) {Gradient \\ estimation};
    \node[rlblock]     (rl)     at (2.50,4.00) {RL Agent};
    \node[criticblock] (critic) at (4.00,3.00)   {Critic};
    \node[simblock]    (sim)    at (4.00,1.75)   {Environment};

    \draw[arrow] (milp.east) -- (bnb.west);
    \draw[arrow] (bnb.east)  -- (policy.west);

    \draw[arrow] (policy.south) -- (grad.north);
    \draw[arrow] (grad.west)
      -- node[pos=0.50,above] {$\nabla_\theta\log\pi_\theta$} (rl.east);

    \draw[arrow] (sim.north) -- (critic.south);
    \draw[arrow] (critic.west) -| (rl.south);
    \draw[arrow, dashed, color={rgb,255:red,0; green,51; blue,102}] 
  (rl.north) -- ($(milp.north)+(-0.2,0.2)$);

    \coordinate (aux1) at ($(sim.west)+(-2.3cm,0)$);
    \draw[arrow,rounded corners=4pt]
      (sim.west)
      -- (aux1)
      -- node[pos=0.50,right] {$s_{t+1}$} (aux1|-milp.west)
      -- (milp.west);

    \coordinate (aux2) at ($(policy.east)+(0.6cm,0)$);
    \draw[arrow,rounded corners=4pt]
      (policy.east)
      -- (aux2)
      -- node[pos=0.50,right] {$a_t$} (aux2|-sim.east)
      --  (sim.east);

  \end{tikzpicture}%
  }
  \caption{Overview of the \textsc{corl} framework}
  \label{fig:overview}
\end{figure}

\section{Background}\label{sec:background}
This section introduces the necessary background on sequential decision making, \ac{rl}, \ac{co}, \ac{milp}, and \ac{bnb}.
\subsection{Modeling Sequential Decision Making Using MDPs}
\ac{mdp} provides a framework to formulate and solve sequential decision-making problems for stochastic systems having the Markovian property \cite{puterman2014markov}. \acp{mdp} consider dynamic systems with underlying  states $ s\in \mathbb S$, actions (decisions) $ a \in \mathbb A$, and the associated stochastic state transition: \(s_+ \sim \mathcal{P} \left(\,.\,|\, s,  a\,\right)\). Solving an \ac{mdp} involves finding a policy $\pi(a| s)$ that minimizes the expected sum of discounted costs $\ell\colon\mathbb S\times\mathbb A\to\mathbb R$ under stochastic closed‐loop trajectories, given as the \ac{mdp} cost: 
\begin{align}\label{eq:MDPCost}
J (\pi) = \mathbb E_{\mathcal{P}^\pi}\left[\left.\sum_{k=0}^\infty \gamma^k \ell\left( s_k, a_k\right)\,  \right |\,  a_k\sim \pi(.| s_k)\right]\,,
\end{align}
for a discount factor $\gamma \in (0,1)$. The expectation is taken over the distribution of \(  s_k \) and \(  a_k \) in the Markov chain induced by $\pi ( a | s)$.  The solution to \ac{mdp} provides an optimal policy $\pi^\star$ from the set $\Pi$ of all admissible policies by minimizing $J(\pi)$, defined as: 
\begin{equation}\label{eq:pi_star}
    \pi^\star = \arg\min_{\pi\in\Pi}\, J\left(\pi\right)\,.
\end{equation}
The solution of an \ac{mdp} is described by Bellman equations~\cite{bellman1957dynamic}:
\begin{subequations}\label{eq:bellman_optimality}
\begin{align}
  Q^\star(s,a) & = \ell(s,a) + \gamma\,\mathbb{E}_{\rho}\bigl[V^\star(s_+)\mid s,a\bigr]\,,\\
  V^\star(s) &= \min_{a} Q^\star(s,a), \\
  \pi^\star(a|s) &= \arg\min_{a} Q^\star(s,a)\,,
\end{align}
\end{subequations}
 where $V^\star$ and $Q^\star$ are the optimal value and action-value functions, respectively. $\mathbb{E}_{{\rho}}$ is the expectation over the distribution of \(\mathcal{P}\). 

\ac{rl} aims to solve the \ac{mdp} approximately, using data. Stochastic policy-gradient is a major class of methods that can be applied to both discrete and continuous state-action space problems. \ac{sac} and \ac{ppo} are two widely used stochastic policy gradient algorithms in \ac{rl} \cite{schulman2017proximal,haarnoja2018soft}. For a differentiable policy $\pi_\theta(a|s)$ parametrized by $\theta$, the  stochastic policy-gradient is:
\begin{equation}\label{eq:spg}
    \nabla_\theta J(\pi_\theta)=\mathbb E_{s,a\sim\pi_\theta}\!\left[\,\nabla_\theta \log \pi_\theta(a|s)\, A^{\pi_\theta}(s,a)\right]\,,
\end{equation}
where $A^{\pi}(s,a)=Q^{\pi}(s,a)-V^{\pi}(s)$ and $Q^{\pi}$, $V^{\pi}$ satisfy the Bellman equations with $\pi$ in place of $\pi^\star$. 

\subsection{Combinatorial Optimization, MILP and B\&B}
\begin{figure}[!t]
  \centering
  \begin{minipage}[t]{0.55\columnwidth}
    \centering
    \resizebox{0.9\columnwidth}{!}{%
          \begin{tikzpicture}[
        every node/.style={font=\scriptsize},
        level 1/.style={sibling distance=3cm, level distance=1.5cm},
        level 2/.style={sibling distance=2cm, level distance=1.5cm},
        normal/.style={circle, draw=black, line width=1pt,
                        minimum size=8mm, inner sep=1pt},
        optimal/.style={circle, draw=green,  line width=1pt,
                        minimum size=8mm, inner sep=1pt},
        pruned/.style={circle, draw=blue, dashed, line width=1pt,
                        minimum size=8mm, inner sep=1pt},
        unexplored/.style={circle, draw=gray, dashed, fill=gray!30,
                          line width=1pt, minimum size=8mm, inner sep=1pt},
        edge from parent/.style={draw,->,line width=0.8pt}
      ]
        \node[normal] (n1) {RN}
          child {
            node[optimal] (n2) {ON}
              edge from parent
          }
          child {
            node[pruned] (n3) {PN}
              edge from parent
            child {
              node[unexplored] (n4) {UN}
                edge from parent
            }
            child {
              node[unexplored] (n5) {UN}
                edge from parent
            }
          };
        \node[font=\scriptsize,right=2pt] at (n2.east) {$Q=-8$};
        \node[font=\scriptsize,right=2pt] at (n3.east) {$Q=-6.3$};
        \node[font=\scriptsize,below left=1pt]  at (n4.south west) {$Q=-6$};
        \node[font=\scriptsize,below right=1pt] at (n5.south east) {$Q=-5$};
      \end{tikzpicture}%
    }
    \vspace{1pt}\newline{(a)}
  \end{minipage}\hfill
  \begin{minipage}[t]{0.40\columnwidth}
    \centering
    \resizebox{0.9\columnwidth}{!}{%
          \begin{tikzpicture}[
        dot/.style={circle, fill=black, inner sep=0pt, minimum size=5pt},
        axis/.style={->, >=stealth, thick, gray!80},
        branch/.style={blue!70, line width=2pt},
        redline/.style={red, line width=2pt},
        mystar/.style={star, star points=5, star point ratio=2.5,
                       fill=yellow, minimum size=12pt, inner sep=0pt}
      ]
        \def\maxval{4}
        \draw[step=1cm, gray!20, thin] (0,0) grid (\maxval,\maxval);
        \draw[axis] (0,0) -- (\maxval+0.5,0) node[anchor=west] {$x$};
        \draw[axis] (0,0) -- (0,\maxval+0.5) node[anchor=south] {$y$};
        \foreach \i in {1,...,\maxval} {
          \draw (\i,-0.1) node[below] {$\i$};
          \draw (-0.1,\i) node[left]  {$\i$};
        }
        \foreach \x in {1,...,\maxval}
          \foreach \y in {1,...,\maxval}
            \node[dot] at (\x,\y) {};
        \draw[branch] (2,0) -- (2,\maxval+0.5);
        \node[anchor=south] at ($(3,4)+(0.2,0.2)$) {$\hat Z$}; 
        \draw[redline] (1.5,0) -- (4,2.5);
        \fill[green] (2,0.5) circle (3pt);
        \node[mystar] at (3,3) {};
      \end{tikzpicture}%
    }
    \vspace{1pt}\newline{(b)}
  \end{minipage}
  \captionsetup{font=small}
  \caption{\textbf{(a)} \ac{bnb} tree: A \ac{bnb} tree may contain unexplored but feasible decisions corresponding to the unexplored leaf nodes (UN). Since their true cost is unknown, we approximate them using their parent, which is a pruned node (PN). \textbf{(b)} Sampling child leaves of a pruned node: For a PN, black dots are unexplored child leaves, the red line represents a constraint, where solutions over it are feasible, and the blue line represents the local branching cut. \(\hat{Z}\) represents the feasible solution space, defined by the blue and red constraints. The green dot is the fractional \ac{lp} solution at the PN, and let the yellow star be the unknown true global optimum for \ac{rl}. The stochastic policy should sample decisions from this space \(\hat{Z}\) for \ac{rl}.}
  \label{fig:bnb_and_sampling}
\end{figure}
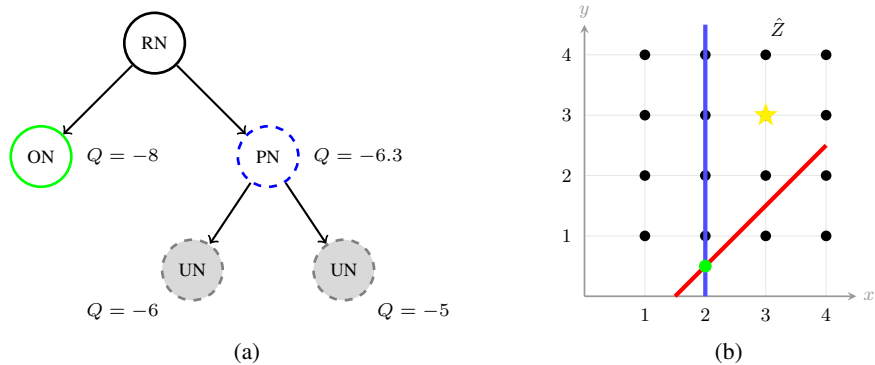

A broad class of \ac{co} problems can be modeled using \ac{milp}.  Typical examples of \acp{milp} include knapsack, facility location, network design, and scheduling \cite{wolsey1999integer}. \acp{milp} are non-convex and NP-hard to solve. The canonical exact algorithm for \ac{milp} is \ac{bnb} \cite{wolsey1999integer}, which solves a series of \ac{lp} relaxations of the  \ac{milp} model. In a \ac{bnb} tree for solving an \ac{milp} (see Fig.~\ref{fig:bnb_and_sampling}a), the \emph{root} node represents the original problem. Each node corresponds to a subproblem obtained by fixing or bounding some variables. Solving its \ac{lp} relaxation produces a lower bound on the objective value \(Q\) of all integer-feasible solutions (unexplored leaf nodes) in the subtree defined by that node. A node is a \emph{leaf} if it is integer-feasible, in which case the bound equals its exact objective value. A processed node that will not be branched further is \emph{fathomed}; this occurs if (i) its relaxation is infeasible, (ii) its bound is no better than the current optimum, or (iii) it is an integer-feasible leaf. A \emph{pruned} node is a fathomed, non-leaf node; pruned nodes provide lower bounds for all their (still unexplored) feasible child leaves, whereas leaf nodes provide exact objective values. 

\section{CORL}\label{sec:method}
In this section, we introduce the \textsc{corl} framework: an actor-critic policy-gradient \ac{rl} approach to learn high-performance combinatorial policies for sequential decision-making as shown in Fig.~\ref{fig:overview}.
First, we introduce the notion of \ac{milp}-based policies for sequential decision making, followed by how \ac{rl} can be used to adapt them for real-world performance, and then we derive a differentiable stochastic \ac{milp} policy that is compatible with \ac{rl}. 

\subsection{MILP Policy for Combinatorial Sequential Decision Making}
When a combinatorial sequential decision‐making problem is subject to uncertainty that resolves over stages, it can be formulated as a multi‐stage \ac{sdp}, wherein iteratively solving a \ac{co} subproblem at each stage \(t\) forms a policy that approximates the optimal solution of the underlying \ac{mdp}~\eqref{eq:MDPCost}. While there are many ways to form such \ac{co}-based policies~\cite{powell2014clearing}, we use a two-stage \emph{parametric} \ac{sdp} formulated as an \ac{milp} and solved using \ac{bnb}:
\begin{subequations}\label{eq:internal_milp}
\begin{align}
\min_{a_t,\;y_t}\quad
  & L_\theta(s_t,a_t,y_t) \;+\; V_\theta(\hat s_{t+1}, y_t)\label{eq:milp_objective}
\\
\text{s.t.}\quad
  & g_\theta(s_t,a_t,y_t) \le 0,
\label{eq:milp_ineq_constraint} \\
  &
    \hat s_{t+1} = h_\theta(s_t,a_t),\quad a_t^i \in \mathbb{Z},\quad i\in\mathcal I\,. \label{eq:milp_eq_constraint}
\end{align}
\end{subequations}
where \(\theta\) represents the \ac{milp} parameters, \(a_t\) represents the decision at stage \(t\), \(L_\theta\) is the instantaneous cost of the state-decision pair.  It may be identical to \(\ell\) in the real-world \ac{mdp}~\eqref{eq:MDPCost}, but it can be different. Function \(g_\theta\) represents the inequality constraints. Function \(h_\theta\) models the transition dynamics of the system, predicting the next state \(\hat{s}_{t+1}\) given the current state \(s_t\) and decision \(a_t\).
For simplicity, we assume $h_\theta(s_t,x_t)$ provides a single next state $\hat{s}_{t+1}$ deterministically.  Ideally, in~\eqref{eq:internal_milp} one would solve a problem along (possibly infinite) future stages; accordingly, the term $V_\theta(\hat s_{t+1})$ serves as a surrogate for the cumulative cost from stage $t\!+\!1$ onward. Ideally, it would equal the optimal value $V^\star(s_{t+1})$~\cite{ruszczynski2003stochastic}; in practice, we approximate this future cost with $V_\theta$. \(y_t\) represents the set of auxiliary variables used to enforce the piecewise linear structure to the functions $L_\theta$, $V_\theta$, and \(g_\theta\). 

Additionally, we assume that the feasibility of decisions is ensured by constraint relaxation, i.e., by penalizing constraint violations in the objective function. In a sequential decision-making context, \eqref{eq:internal_milp} defines a policy: at any given stage \(t\), it solves the two-stage optimization given the current state \(s_t\) to derive model-based decisions \(a_t\) that minimize the given objective. This two-stage formulation~\eqref{eq:internal_milp} simplifies solving the multi-stage problem: instead of optimizing across all the stages, we solve a single-stage \ac{milp} augmented with the value approximation $V_\theta$. However, this formulation naturally extends to the full multi-stage setting.

Assuming the \ac{milp} scheme~\eqref{eq:internal_milp} has a unique minimizer,  we can view the \ac{milp} scheme~\eqref{eq:internal_milp} as a model of the action-value function \(Q\) in Bellman equations~\eqref{eq:bellman_optimality} by fixing the decision variable \(a_t\) in \eqref{eq:internal_milp}:
\begin{equation}
\begin{aligned}\label{eq:Q_model_milp}
Q_\theta(s_t,\hat{a}_t)
:= \;&\eqref{eq:milp_objective}
\\
\text{s.t. } &\eqref{eq:milp_ineq_constraint}\;\text{--}\;\eqref{eq:milp_eq_constraint}\,,\quad a_t=\hat{a}_t\,.
\end{aligned}
\end{equation}
We use the term objective value or action-value for \(Q_\theta\) interchangeably. This definition of $Q_\theta$ is valid in the sense of fundamental Bellman relationships~\eqref{eq:bellman_optimality} between optimal action-value functions, value functions, and policies, i.e. 
\begin{subequations}\label{eq:bellman_milp}
\begin{align}
V_\theta\left(s\right) &= \min_{ a}\, Q_\theta\left(s, a\right),\\\quad 
\pi_\theta\left(s\right) & = \arg\min_{a}\, Q_\theta\left(s,a\right)\,.\label{eq:milp_policy_bellman}
\end{align}
\end{subequations}
In this sense, the \ac{milp} model~\eqref{eq:internal_milp} can be seen as a model of the real-world \ac{mdp}. This connection helps to explain the theoretical basis of the \textsc{corl} framework developed in the following sections.

\subsection{CORL Problem Formulation}

The overarching goal of any \ac{co}-based decision-making scheme is to derive optimal decisions for the real-world system. Although accurately modeling a stochastic real-world system with an \ac{milp} is nearly impossible, it has been shown that such inexact decision models can nonetheless recover the true optimal policy \(\pi^\star\)~\cite{anand2025all}. This demands the \ac{milp} policy~\eqref{eq:milp_policy_bellman}  to match the true optimal policy for the real system, \(\pi^\star\)~\eqref{eq:pi_star}. Assuming the model of \(Q_\theta\) given by~\eqref{eq:Q_model_milp} is bounded and the \ac{milp} parameterization is rich enough, this (necessary and sufficient) optimality condition can be represented as:
\begin{align}
\label{eq:optimality_condition}
\arg\min_{a}\,{Q}_\theta\left( s,a\right)=\arg\min_{ a}\,Q^\star\left(s, a\right),\quad \forall\,s \,.
\end{align}

The identification of the optimal \ac{milp} parameters~$\theta^\star$ that satisfy the optimality condition~\eqref{eq:optimality_condition} can be framed as the following optimization problem: 
\begin{equation}\label{eq:rl_over_milp}
\theta^\star \;\in\; \arg\min_{\theta}\; J(\pi_\theta)\,,
\end{equation}
where \(J\) is given by~\eqref{eq:MDPCost} and the policy \(\pi_\theta\) is given by the solution to~\eqref{eq:internal_milp}.  The core principle of our \textsc{corl} framework is to \emph{employ \ac{rl} to approximately solve this optimization problem~\eqref{eq:rl_over_milp} directly from real-world data.} In \textsc{corl}, the policy consists of the \ac{milp} model~\eqref{eq:internal_milp} and the \ac{bnb} solver, where the \ac{rl} agent adapts the \ac{milp} parameters \(\theta\) to maximize real‐world performance (see Fig.~\ref{fig:overview}). Within \ac{rl}, stochastic policy-gradient methods provide a robust framework for learning stochastic policies~\cite{haarnoja2018soft}. We adopt stochastic policy-gradient methods in an actor–critic architecture to solve the \textsc{corl} problem (see Fig.~\ref{fig:overview}): the actor implements the \ac{milp}-based policy, while the critic can be implemented using a generic function approximator such as neural networks.  The stochastic policy-gradient methods require the policy to be both stochastic and differentiable, properties that the \ac{milp} policy~\eqref{eq:internal_milp} lacks. To address this, in the next two sections, we derive a differentiable stochastic policy from this discrete \ac{milp} policy.

\subsection{Deriving a Stochastic Policy from B\&B}
We can derive a differentiable stochastic policy by applying a softmax distribution over the \(Q_\theta\)-values of the \ac{milp} scheme~\eqref{eq:Q_model_milp}:
\begin{equation}
  \pi_\theta(a_t | s_t)
  = \frac{e^{-\beta Q_\theta(s_t,a_t)}}
         {\sum_{\tilde a_t\in\mathcal{A}} e^{-\beta Q_\theta(s_t,\tilde a_t)}} ,
  \label{eq:naive_pol}
\end{equation}
where $\beta>0$ represents the exploration parameter. However, directly applying \eqref{eq:naive_pol} is infeasible because \ac{bnb} algorithms do not explore every feasible decision $a_t\in\mathcal{A}$ (leaf), so their true objective values $Q_\theta(s_t,a_t)$  are not known. Commercial solvers, e.g.~\cite{gurobi}, can return a subset of feasible decisions with their objective values, which can be used to form a policy. However, that would exclude a possibly large set of feasible decisions, which would be beneficial to explore (see Fig.~\ref{fig:bnb_and_sampling}b). Therefore, we exploit the structure of the \ac{bnb} tree to derive an alternative form to the policy~\eqref{eq:naive_pol}.

Let $\mathcal{K}_t$ be the set of all explored nodes $k$ (leaf or pruned) in the \ac{bnb} tree at \(t\). All such (leaf or pruned) nodes \({k \in \mathcal{K}_t}\) is assigned a scalar objective value that we denote by $Q_\theta^{k,t}$. For a pruned node, $Q_\theta^{k,t}$ is a lower bound on every (still unexplored) child nodes, and for a leaf, it is the exact objective of its integer‐feasible solution. Formally,
\begin{equation}\label{eq:node_bound_relation}
  Q_\theta(s_t,a) \;\ge\; Q_\theta^{k,t}\quad \forall a\in\mathcal{A}_k\,,
\end{equation}
with \(Q_\theta(s_t,a_{t,k}) \;=\; Q_\theta^{k,t}\) if $k$ is a leaf. $\mathcal{A}_k$ denotes the set of feasible decisions at node $k$, i.e.\ the unexplored child nodes belonging to $k$, and let $a_{t,k}$ denote the integer solution stored at an already‐explored leaf. 

We can use \(Q_\theta^{k,t}\) to sample a node \(k\) by a softmax over the finite set $\mathcal{K}_t$ as:
\begin{equation}
  P(k\mid s_t)
  = \frac{e^{-\beta Q_\theta^{k,t}}}
         {\sum_{i\in\mathcal{K}_t} e^{-\beta Q_\theta^i}} .
  \label{eq:node_prob}
\end{equation}

Sampling a node $k$ from~\eqref{eq:node_prob} allocates probability mass both to explored leaves and to promising pruned nodes through their \(Q_\theta^{k,t}\) values. This allows us to score unexplored regions of the tree and bias exploration toward branches with a better lower bound. Since for a pruned node, \(Q_\theta^{k,t}\) is only a lower bound, their $P(k| s_t)$ in~\eqref{eq:node_prob} can be \emph{optimistic}, as a pruned branch may receive higher probability than it would if exact values of its children were available (see.~\eqref{eq:naive_pol} and Fig.~\ref{fig:bnb_and_sampling}a). At the same time, a pruned node with a poor \(Q_\theta^{k,t}\) may contain a large number of child leaves, which may not be reflected in the distribution~\eqref{eq:node_prob}. However, for the proof-of-concept, we limit our methodology to~\eqref{eq:node_prob}. 

Now, in order to approximate the intractable softmax over $\mathcal{A}$ in~\eqref{eq:naive_pol}, a decision $a_t$ ought to be selected randomly among the child leaf nodes of a node \(k\) sampled according to~\eqref{eq:node_prob}. To that end, let us define the optimal solution (integer or fractional) at a (pruned or leaf) node \(k\) corresponding to \(Q^{k,t}_\theta\) as \(a^\star_{k,t}\). A decision \(a_t\) can be sampled from a selected node $k$ as follows:
\begin{itemize}
  \item {If \(k\) is a leaf node:} we take its integer decision \({a_t = a^\star_{k,t}}\) deterministically.
  \item {If \(k\) is a pruned node:} we sample an (unexplored) child leaf node using a heuristic (see Fig.~\ref{fig:bnb_and_sampling}b). For the proof-of-concept, we propose two simple heuristics:\\  
    (1)~\emph{Uniform sampling}: treat all integer-feasible decisions \(a_t \in \mathcal{A}_t\) corresponding to the (unexplored) child leaves of the node $k$ as equally likely, assuming it has a finite set of decisions and they are known.\\  
    (2)~\emph{\ac{nns}}: sample a decision \(a_t\) from the neighbourhood of the fractional solution $a^\star_{k,t}$ of the pruned node. This can be represented using a softmax over (negative) Manhattan distances with \(\beta_d\) as the temperature parameter:
    \begin{equation}
  p(a_t\mid k,s_t)
  \;\propto\;
  e^{-\beta_d\,\|a_t - a_{t,k}^\star\|}\,.
\end{equation}
\end{itemize}

Note that by construction, the \ac{bnb} tree induces a partition in which each integer–feasible decision~$a_t$ belongs to a unique (leaf or pruned) node \(k^\star\!\in\!\mathcal{K}_t\). Therefore, only that particular node \(k^\star\) contributes to the probability of its associated decision $a_t$ in the policy. 

Putting these together forms an approximation to the \ac{milp} policy as:
\begin{equation}\label{eq:stochastic_policy}
      \pi_\theta(a_t | s_t) \;=\; P(k^\star\mid s_t)\;\; p(a_t\mid k^\star,s_t),
\end{equation}
where $p(a_t\mid k^\star,s_t)=1$ if $k^\star$ is a leaf.  For a pruned node, $p(\cdot\mid k^\star,s_t)$ is given by a sampling strategy (e.g., uniform sampling or \ac{nns} heuristic).

\subsection{Policy Gradient Estimation}
The term $p(a_t| k^\star,s_t)$ in~\eqref{eq:stochastic_policy} depends on $\theta$ via the cardinality of node $k^\star$, i.e., the number of its child leaf nodes. The cardinality can change when $\theta$ alters the pruning pattern. However, the cardinality of a node \(k^\star\) is piecewise constant in \(\theta\). Therefore, the gradient built by assuming it as a constant is correct almost everywhere except on the zero-measure where the cardinality changes. Therefore, we assume \(\nabla_\theta \log p(a_t\mid k^\star,s_t)\) is zero when it is defined. As a result, the gradient of the log policy needed for stochastic policy gradient~\eqref{eq:spg} follows from~\eqref{eq:stochastic_policy}:
\begin{equation}\label{eq:log_grad}
\begin{aligned}
    \nabla_\theta \log \pi_\theta(a_t | s_t) &= \nabla_\theta \log P(k^\star \mid s_t) \\
    &=  - \beta \nabla_\theta Q^{k^\star,t}_\theta -\nabla_\theta \log \sum_{i \in \mathcal{K}} e^{-\beta Q^{k,t}_\theta}\,. 
\end{aligned}
\end{equation}

In order to calculate \(\nabla_\theta  Q^{k,t}_\theta\) at any node \(k\), we consider the partial Lagrangian of the \ac{milp} scheme~\eqref{eq:internal_milp} given by: 
\begin{equation}\label{eq:lagrangian}
\begin{aligned}
\mathcal L(a_t,y,\lambda,\mu;\theta) 
= L_\theta(s_t,a_t,y_t) &+ V_\theta(\hat s_{t+1},y_t)\\
 &+ \lambda^\top g_\theta(s_t,a_t,y_t)\\
 &+ \mu^\top\bigl(\hat s_{t+1} - h_\theta(s_t,a_t)\bigr)\,.
\end{aligned}
\end{equation}
where $\lambda$ and \(\mu\) represents the vector of Lagrange multipliers for inequality and equality constraints, respectively. At a pruned node, the \ac{kkt} stationarity condition
$\nabla_{a^\star_t}\mathcal L=0$ holds, so the total derivative of $\mathcal L$ with respect to $\theta$ reduces to its partial derivative.
At a leaf node, the solution $a_t^\star$ is piecewise constant in $\theta$, hence ${\nabla_\theta a_t^\star=0}$ almost everywhere. In both cases (at a leaf or pruned node),
$\nabla_\theta Q_\theta$ is obtained by differentiating the Lagrangian
with respect to $\theta$ while holding the optimal primal/dual variables fixed. Therefore, under standard regularity conditions, the envelope theorem~\cite{milgrom2002envelope}
implies that, at the optimal solution \((a^\star_t,,y^\star,\lambda^\star,\mu^\star)\):
\begin{equation}
      \nabla_\theta Q_\theta(s_t,a^\star_t)
  = \left.\nabla_\theta \mathcal L\right|_{(a^\star_t,y^\star,\lambda^\star,\mu^\star)}.
  \label{eq:envelope}
\end{equation}
 Since both $L_\theta$ and $V_\theta$ are piecewise‐linear in $\theta$, the auxiliary variables $y$ act as linear coefficients that remain constant within each region, so $\nabla_\theta y^\star=0$ almost everywhere; resulting in:  
 \begin{equation}\label{eq:milp-gradient}
\begin{aligned}
  \nabla_\theta Q_\theta(s_t,a_{t,k}^\star)
  = \nabla_\theta L_\theta(s_t,a_{t,k}^\star)
   &+ \nabla_\theta V_\theta(\hat s_{t+1})\\
  &+ \bigl(\lambda^\star\bigr)^\top \nabla_\theta g_\theta(s_t,a_{t,k}^\star)\\
   &- \bigl(\mu^\star\bigr)^\top \nabla_\theta h_\theta(s_t,a_{t,k}^\star)\,.
\end{aligned}
 \end{equation}
Note that \(a^\star_{k,t}\) is the optimal solution (integer or fractional) at a (pruned or leaf) node \(k\) corresponding to \(Q^{k,t}_\theta\). Since each  \ac{bnb} subproblem is solved to optimality,  the optimal primal solution (\(a^\star_{t,k}\)) and dual solution (\(\lambda^\star\, \mu^\star \)) of the \ac{milp} subproblem at node \(k\) satisfy the \ac{kkt} conditions~\cite{kuhn2013nonlinear} and ensure that \eqref{eq:milp-gradient} holds exactly.
Note that, \eqref{eq:milp-gradient} assumes that any discontinuity caused by a change in the active integer solution arises only at a measure-zero set and therefore does not affect the gradient-based update in practice. 

An illustrative pseudocode for a basic on‐policy \ac{rl} implementation of the \textsc{corl} methodology is provided in Algorithm~\ref{alg:corl}.
\begin{algorithm}[H]
\caption{}\label{alg:corl}
\begin{algorithmic}[1]
    \State \textbf{Input:} Parametric MILP‐actor $\theta$, critic $\phi$
    \FOR{iter = 1 to $N$}
        \State Initialize buffers: $\mathcal{Q}\leftarrow\{\}$,\quad 
        $\mathcal{D}_{PG}\leftarrow\{\}$,\quad 
        $\mathcal{D}\leftarrow\{\}$
        \FOR{$t=1$ to $T$}
            \State Solve \ac{milp} with \ac{bnb}, append $(Q^{k,t}_\theta, a^\star_{k,t})$ to $\mathcal{Q}$
            \State Form $P(k\mid s_t)$ via \eqref{eq:node_prob} over $\{Q_\theta^k\}$
            \State Sample node $k\sim P(\cdot\mid s_t)$
            \IF{$k$ is leaf}
                \State $a_t \leftarrow a^\star_{k,t}$
            \ELSE
                \State $a_t\sim p(a\mid k,s_t)$ \Comment{uniform or \ac{nns}}
            \ENDIF
            \State Compute $\nabla_\theta Q_\theta(s_t,a^\star_{k,t})$ via \eqref{eq:milp-gradient}
            \State Compute $\nabla_\theta\log\pi_\theta$ via \eqref{eq:log_grad} and append to $\mathcal{D}_{PG}$
            \State Apply $a_t$ to the environment
            \State Append $(s_t,a_t,r_t,s_{t+1})$ to $\mathcal{D}$
        \ENDFOR
        \State Update critic \(\phi\) on $\mathcal{D}$
        \State Evaluate policy on $\mathcal{D}$, compute advantages $A$
        \State Update actor using $\mathcal{D},\mathcal{D}_{PG},A$ via \eqref{eq:spg}
    \ENDFOR
\end{algorithmic}
\end{algorithm}

\section{Illustrative Examples}\label{sec:examples}
As an illustrative example for \textsc{corl}, we consider a \ac{co} problem governed by a linear stochastic transition function formulated as:
\begin{equation}\label{eq:ex_prob}
    \begin{aligned}
        \min\limits_{a_t} \quad & \sum_{t=1}^T \ell^\intercal a_t + p\zeta_t\\
        \text{s. t.,}  \quad & D s_t + Ea_t \leq F  + \zeta_t\,,\\
        & s_{t+1} = Ms_t+Ba_t +w_t\,, \\
        & lb\leq a_t \leq ub\,, \quad \zeta_t \geq 0\,, \quad a^i_t \in \mathbb{Z}\,, \quad \forall a_t^i \in a_t\, \quad t  =1,\dots,T\,.
    \end{aligned}
\end{equation}
Where \(\ell\) is the cost of a decision \(a_t\), \(s_t\) is the state at stage \(t\). The matrices \(D,E,F\) define any set of linear inequality constraints, modeling the known constraints of the problem. \(M,B\) define the linear dynamics of the system, while \(w_t \sim \mathcal{N}(0,\sigma^2)\) represents a normally distributed disturbance affecting the system transition. All entries in \(a_t\) are integers, while \(s_t\) can be continuous. \(\zeta\) is a continuous slack variable that penalizes violation of the inequality constraint. The magnitude of the penalization is set to be proportional to the deviation from the constraints with a factor \(p\). The terms \(lb\) and \(ub\) are the lower and upper bounds on \(a_t\).

We formulate an \ac{milp} model of \eqref{eq:ex_prob} that represents a single stage \(t\) and summarizes the remaining stages in a value function, forming a two-stage \ac{milp}:
\begin{equation}
    \begin{aligned}        
        \min\limits_{a_t,v,\zeta_t} \quad & L^\intercal a_t + v + p\zeta_t\\ 
        \text{s. t.,} \quad & D s_t + E a_t  \leq F + \zeta_t \,, \\
        & \psi_j (\hat{M} s_t + \hat{B} a_t ) + b_j \leq v\,, \\
        &lb\leq a_t \leq ub\,, \quad \zeta_t \geq 0\,, \quad a^i_t \in \mathbb{Z}\,, \quad \forall a_t^i \in a_t\,. \\
    \end{aligned}
    \label{eq:milp_model}
\end{equation}

Here \(v\) denotes the value function estimate, defined as a piecewise linear function of the next state  \(s_{t+1}\). Since the value function of a stochastic program is a convex polyhedral function, we estimated it with a set of affine value function pieces \(\psi_j,b_j\).  \(D\), \(E\), \(F\), \(lb\) and \(ub\) are assumed to be known. The state transition and immediate cost are assumed to be unknown, and \(L,\hat{M},\hat{B}\) denotes their estimates. The value function pieces \(\psi_j,b_j\) are to be learned. However, to simplify gradient calculation, the products of \(\psi_j\hat{M}\) and \(\psi_j\hat{B}\) are learned instead of learning \(\psi_j,\hat{M},\hat{B}\) separately. We collect all the parameters of the model that need to be learned into \(\theta = \{L, \psi_j\hat{M},\psi_j\hat{B},b_j\}\).

\subsection{Gradient Estimation}The Lagrangian for a state-decision pair is computed as: 
\begin{equation}\label{eq:ex_lagr}
\begin{aligned}
         \mathcal{L}(s_t, a_{t},\lambda_t) & =  \lambda_{ineq}(F + \zeta_t - Ds_t-Ea_{t})\\
         & \qquad + \sum_j\lambda_j(v-b_j-\psi_j\hat{M}s_t-\psi_j\hat{B}a_{t})\\
        & \qquad + \lambda_{lb}(a_{t}-lb) + \lambda_{ub}(ub-a_{t}) +L^\intercal a_{t} + v + p\zeta_t\,. 
    \end{aligned}
\end{equation}
Where \(\lambda_t = [\lambda_{ineq},\lambda_1,\dots ,\lambda_J,\lambda_{lb},\lambda_{ub}]^\intercal \),  each \(lb\) and \(ub\) are specific to the subproblem \(j\) from \(a_t\) was sampled. We can then compute the following:
\begin{equation}
   \bigg[ \frac{\partial\mathcal{L}}{\partial L} \,,\,\frac{\partial\mathcal{L}}{\partial \psi_j\hat{M}}\,,\,\frac{\partial\mathcal{L}}{\partial \psi_j\hat{B}}\,,\,\frac{\partial\mathcal{L}}{\partial b_j}\bigg] \;= \; \bigg[ a_{t}\,,\, - \lambda_js_t\,,\, -\lambda_ja_{t}\,,\, -\lambda_j\bigg]\,.
\end{equation}
The gradient of \(Q_{\theta}(s_t,a_t)\) is given by \eqref{eq:envelope} as:
\begin{equation}
    \nabla_\theta Q_{\theta}(s_t,a_t) = 
\bigg[\frac{\partial\mathcal{L}}{\partial L} \,,\,\frac{\partial\mathcal{L}}{\partial \psi_j\hat{M}}\,,\,\frac{\partial\mathcal{L}}{\partial \psi_j\hat{B}}\,,\,\frac{\partial\mathcal{L}}{\partial b_j}\bigg]\,.
\end{equation}

\begin{table}[t]
  \centering
  \setlength{\tabcolsep}{4pt}
  \renewcommand{\arraystretch}{1.05}
  \begin{tabular}{@{}c c c@{}}
    {Variable} & {\(\quad ub \quad\)} & {\(\quad lb \quad\)} \\
    \midrule
    $\ell,\;L$                                & 10   & 0   \\
    $D_1,\;E,\;B$                          & 1    & 0   \\
    $F_1$                                   & 15   & 5   \\
    $F_2$                                   & 10   & 1   \\
    $M,\;\psi_j\hat M,\;\psi_j\hat B,\;b_j$& 0.1  & 0   \\
  \end{tabular}
  \caption{Summary of uniform sampling upper and lower bounds for all problem parameters used in the experiments. Fixed parameters: $D_2=0$, $a_{lb}=0$, $a_{ub}=10$.  ${D=[D_1;D_2]}$ and ${F=[F_1;F_2]}$.}
  \label{tab:experiment details}
\end{table}


\begin{figure*}[!t]
  \centering
  \resizebox{1\columnwidth}{!}{\input{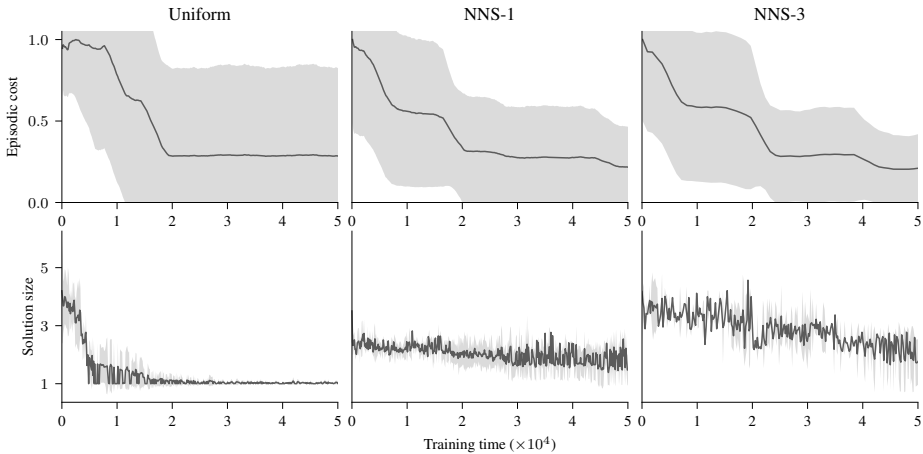}}
  \captionsetup{font=small}
  \caption{(\textbf{top row}): Mean and variance of the training progress over 5 random seeds for the example problem \eqref{eq:ex_prob}. Each subplot corresponds to a different sampling method (i) uniform sampling, (ii) \ac{nns} using 1 nearest neighbour  (NNS-1), and (iii)  \ac{nns} using 3 nearest neighbours (NNS-3). The episodic cost is smoothed using a rolling window of width \(W = 10\) and its values are normalized between 0 and 1. (\textbf{bottom row}): Comparison on how the size of the solution set changes during training for each of the sampling methods.  The size of the solution set is smoothed using a rolling window of width \(W = 10\).}
  \label{fig:exp2_and_3}
\end{figure*}

\subsection{RL Experiments}
All \ac{rl} experiments are conducted on the example in \eqref{eq:ex_prob}. We treat reward as (negative) cost, so the objective is to minimize the discounted cumulative cost.
For all experiments, the matrix and vector variables defining the problem~\eqref{eq:ex_prob} are randomly generated with different seeds (see Table~\ref{tab:experiment details}). 
In all experiments, we set the critic learning rate at 0.01 and the actor learning rate at 0.001.  The penalty factor \(p\) in \eqref{eq:ex_prob} was set to 1000, the discount factor \(\gamma\) for the \ac{td} critic to 0.9, and the exploration‐noise standard deviation \(\sigma\) to 1. For all \ac{rl} experiments, we used a stochastic policy‐gradient setup in which the \ac{milp} model~\eqref{eq:milp_model} and a \ac{bnb} solver form the actor, and an \ac{nn}-based \ac{gae} serves as the critic. The policy gradient is calculated using \eqref{eq:log_grad}. We conducted two sets of experiments. In all the experiments, both \(L\) and \(v\) are learned using \ac{rl}. Both experiments were repeated five times with different random seeds.

The objective of Experiment~(i) is to demonstrate that the \textsc{corl} algorithm can both learn from data and converge while additionally comparing the effect of the two sampling methods used. For each seed, two training runs were performed, one with uniform sampling and one with a \ac{nns}. For \ac{nns}, we evaluate two variants: NNS-1 (single nearest neighbor) and NNS-3 (3 nearest neighbors), where candidates are chosen by shortest Manhattan distances to the fractional solution at the pruned node.  The results are shown in the top row of Fig.~\ref{fig:exp2_and_3}. We observe that the cost curves are decreasing and converging, demonstrating the effectiveness of the algorithm. This implies that the agent can improve its \ac{milp} policy by \ac{rl}-based fine-tuning of its parameters. We additionally note that both the \ac{nns} and the uniform sampling approaches tend to converge to similar episodic costs. However, the \ac{nns} approach seems to perform better, is more robust across different seeds, and tends to perform better early in the training. The sudden jumps in the learning curve are natural due to the discontinuous nature of \acp{milp}. 

Experiment (ii) monitor how the size of the solution set (number of leaves and pruned nodes) in the \ac{bnb} solver changes throughout training. The bottom row of Fig.~\ref{fig:exp2_and_3} shows the results, where for all 5 seeds the size of the solution set shrinks consistently, regardless of the sampling method used. In this particular example, we observe that learning converges when the size of the solution set collapses to a single element. Since the size of the solution set is directly linked to the \ac{bnb} tree size, the size of the \ac{bnb} tree also shrinks as during convergence. NNS-3 appears to be the most explorative of the three, as its solution set shrinks more gradually compared to the others.

\section{Discussion}\label{sec:discussion}
Results in Section~\ref{sec:examples} validate the \textsc{corl} proof-of-concept, demonstrating that a general \ac{milp} model can be fine-tuned via \ac{rl} on real-world data to directly optimize its decision performance. Importantly, rather than trying to accurately model the real‐world problem in the traditional sense, the \textsc{corl} approach uses \ac{rl} to adapt a \ac{co} decision scheme directly for maximizing its real-world performance while preserving its \ac{milp} structure~\cite{anand2025all}. While this work provides the mathematical foundations and a proof-of-concept, multiple directions of further work are needed to realize its potential in real-world applications. From the \ac{rl} perspective, developing scalable \textsc{corl} algorithms based on \ac{sac} and \ac{ppo} would be highly valuable. Additionally, integrating offline \ac{rl} methods into \textsc{corl} is essential for practical, data-efficient deployment. From the \ac{co} side, the proof-of-concept should be extended to integrate with commercial \ac{bnb} solvers, improving sampling strategies, and further exploiting the structure of the \ac{co} problem to enhance learning are interesting lines of future work. Solving an \ac{milp} problem at every \ac{rl} step can be computationally prohibitive for large problems. In practice, this motivates using solver acceleration such as warm-starting, tree reuse, heuristics, or applying multiple policy-gradient updates using local or learned policy approximations.  Additionally, \textsc{corl} needs to be extended beyond the specific \ac{milp}+\ac{bnb} combination to a broader class of \ac{co} methods.

Regarding sampling the nodes, using~\eqref{eq:node_prob} is effective as a proof-of-concept; however, it suffers from two biases. On one hand, it can be overly optimistic, assigning high probability to pruned branches; and on the other hand, overly pessimistic, since a low \(Q_\theta^{k,t}\) may mask a large number of potential child leaves. Addressing these biases is an important direction for future work. We find that the decision sampling strategy strongly affects \ac{rl} convergence. The \ac{nns} outperforms uniform sampling: it converges faster, is less noisy, and achieves higher returns early on. Unlike uniform sampling, \ac{nns} leverages pruned‐node solutions and branch bounds to focus exploration on high‐value regions, reducing suboptimal action selection and smoothing the learning. However, \ac{nns} could considerably reduce exploration when the \ac{bnb} solution set shrinks, since each pruned node contributes only a few representative decisions. Therefore, the choice of how many neighbors to include, more broadly, how to balance exploration versus exploitation, remains problem-specific. As training progresses, \ac{rl} could lower the objective values of suboptimal branches, prompting their early pruning and shrinking the \ac{bnb} tree. As the \ac{bnb} solution set collapses toward convergence, the policy gradient approaches zero, as demonstrated in the example. 

In our analysis, we assume all decisions are feasible in order to restrict the scope of this work. However, for most real-world problems, relaxing a constraint by penalizing the cost might not be applicable. Therefore, a key direction for future work is needed to address this limitation by deriving sampling schemes that guarantee feasibility. Additionally, calculating the policy gradient over a discontinuous policy function can be problematic, since the true gradient is undefined at the discontinuities. The Lagrangian of~\eqref{eq:internal_milp}, changes instantaneously as constraints activate or deactivate, and the approximate gradient in \eqref{eq:envelope} is computed without accounting for these discontinuities. Similarly, we ignore the dependence of the cardinality of the feasible solution set on policy parameters~\(\theta\). Although this may produce non-zero gradients at discontinuities, such events are rare, and their errors tend to average out; nonetheless, further work is needed to understand the implications of these simplifications and to devise strategies for explicitly handling such discontinuities. Results from our illustrative example demonstrate the effectiveness of the \textsc{corl} approach despite these approximations; however, evaluation on real-world problems is necessary to confirm its robustness, and we will pursue this in future work.

On an interesting side note, the proposed \textsc{corl} framework demonstrates how an \ac{milp} scheme can serve as a compatible function approximator for \ac{rl}, extending mathematical programming–based structured policies for \ac{rl}~\cite{reiter2025synthesis}. By embedding an \ac{milp} policy, the agent can exploit the problem structure more effectively and reduce the sample complexity compared to using mode-free \ac{rl}. We further hypothesize that \textsc{corl} could enable the development of safe and explainable \ac{rl} policies for combinatorial decision-making problems.

\section{Conclusions}\label{sec:conclusions}
We presented \textsc{corl}, a proof-of-concept \ac{rl} framework for combinatorial decision-making specific to \acp{milp} with \ac{bnb} solver. We formulated the \ac{milp} with a \ac{bnb} solver as a stochastic policy, and derived its stochastic policy gradient to make the \ac{milp} scheme a compatible policy for \ac{rl}. Compared to the traditional approach of improving the modeling accuracy of \ac{milp} to enhance decision quality, the \textsc{corl} framework directly fine‐tunes an existing \ac{milp} scheme to maximize its real-world performance through \ac{rl}. We evaluated the \textsc{corl} framework over an illustrative combinatorial decision-making problem. The \textsc{corl} framework provides the necessary preliminaries for building a scalable \textsc{corl} algorithm for learning high-performance combinatorial policies for real-world processes. However, a broad range of future directions needs to be explored thoroughly to realize its potential real-world applicability.

\bibliographystyle{splncs04}
\bibliography{mbrl}

\end{document}